\documentclass[10pt,twocolumn,letterpaper]{article}

\usepackage{cvpr}
\usepackage{times}
\usepackage{epsfig}
\usepackage{graphicx}
\usepackage{amsmath}
\usepackage{amssymb}
\usepackage{booktabs}
\usepackage{multirow}
\usepackage{color}
\usepackage{array,ragged2e}
\usepackage{subfig}
\usepackage{placeins}

\usepackage{lipsum}

\usepackage[font=small,skip=0pt]{caption}
\usepackage[breaklinks=true,bookmarks=false]{hyperref}

 \makeatletter
 \DeclareRobustCommand\onedot{\futurelet\@let@token\@onedot}
 \def\@onedot{\ifx\@let@token.\else.\null\fi\xspace}

 \makeatother

\newcommand{\camera}[1]{\textcolor{black}{#1}}
\newcommand{\modelName}[1]{DCC}

\newcommand{\myparagraph}[1]{\noindent \textbf{#1}.\ }

\cvprfinalcopy %

\def\cvprPaperID{1600} %

\ifcvprfinal\fi
\begin{document}

\title{Deep Compositional Captioning: \\ Describing Novel Object Categories without Paired Training Data}

\author{Lisa Anne Hendricks$^1$\\
\and
Subhashini Venugopalan$^3$\\
\and
Marcus Rohrbach$^{1,2}$\\
\and
Raymond Mooney$^3$\\
\and
Kate Saenko$^4$\\
\and
Trevor Darrell$^{1,2}$\\
\and
$^1$ UC Berkeley,\  \ $^2$ ICSI, Berkeley,\  \ $^3$ UT Austin,\  \ $^4$ UMass Lowell\\
}

\maketitle
\begin{abstract}

While recent deep neural network models have achieved promising results on the image captioning task, they rely largely on the availability of corpora with paired image and sentence captions to describe objects in context. In this work, we propose the Deep Compositional Captioner (\modelName{}) to address the task of generating descriptions of novel objects which are not present in paired image-sentence datasets. 
Our method achieves this by leveraging large object recognition datasets and external text corpora and by transferring knowledge between semantically similar concepts.
Current deep caption models can only describe objects contained in paired image-sentence corpora, despite the fact that they are pre-trained with large object recognition datasets, namely ImageNet.  In contrast, our model can compose sentences that describe novel objects and their interactions with other objects. We demonstrate our model's ability to describe novel concepts by empirically evaluating its performance on MSCOCO and show qualitative results on ImageNet images of objects for which no paired image-sentence data exist. Further, we extend our approach to generate descriptions of objects in video clips. Our results show that \modelName{} has distinct advantages over existing image and video captioning approaches for generating descriptions of new objects in context. 

\end{abstract}

\section{Introduction}
\begin{figure}[t]
\begin{center}
\includegraphics[width=1\linewidth]{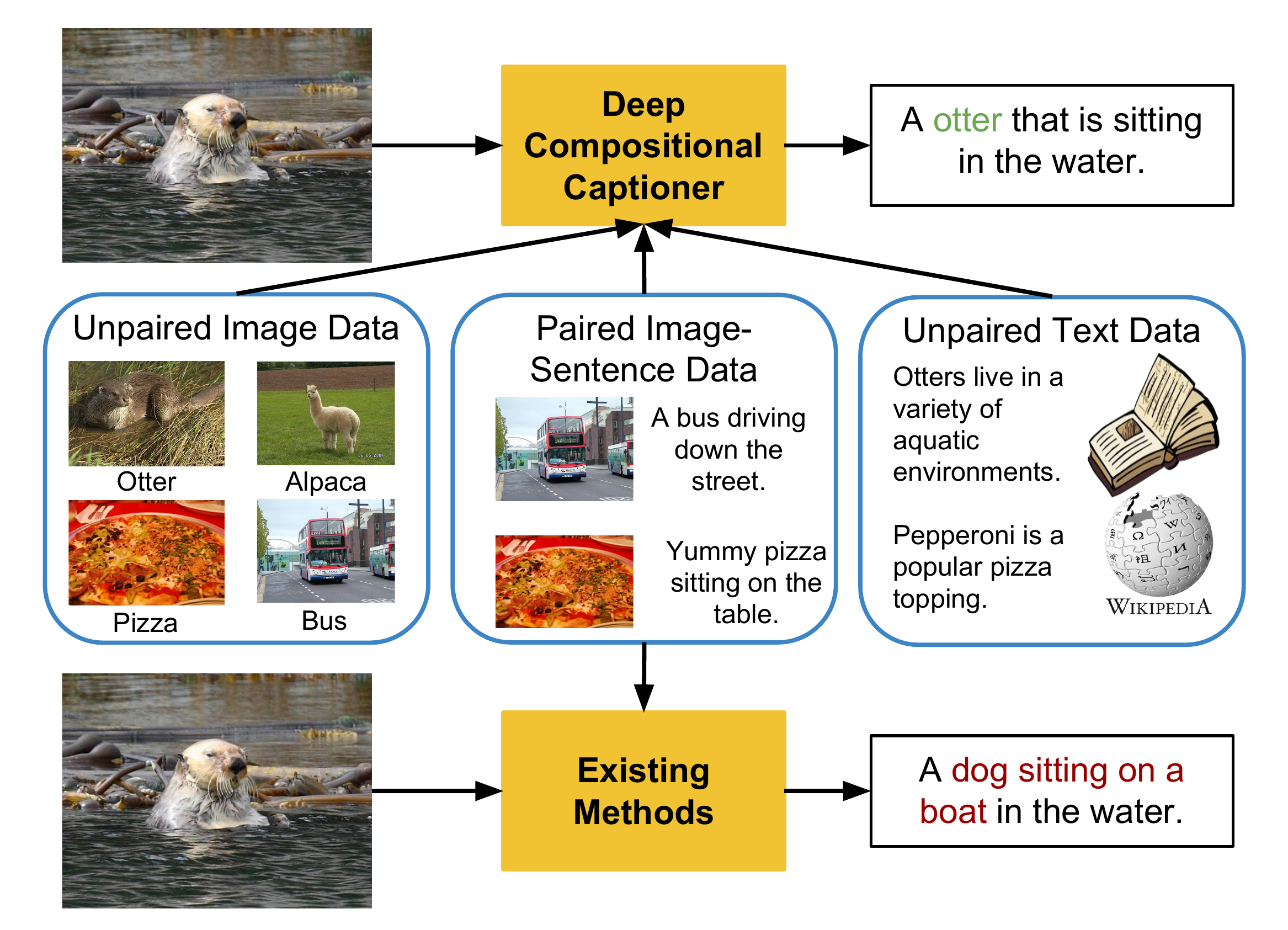}
\end{center}
   \caption{Existing deep caption methods are unable to generate sentences about objects  unseen in caption corpora (like otter).  In contrast, our model effectively incorporates information from independent image datasets and text corpora to compose descriptions about novel objects without any paired image-sentence data.}
\label{fig:teaser}
\end{figure}

In the past year, several deep recurrent neural network models have demonstrated promising results on the task of generating descriptions for images and videos \cite{vinyals14arxiv,donahue15cvpr,kirosArxiv14,karpathy14arxiv,mao15iccv}. Large corpora of paired images and descriptions, such as MSCOCO~\cite{coco2014} and Flickr30k~\cite{flickr30k} have been an important factor contributing to the success of these methods.  However, these datasets describe a relatively small variety of objects in comparison to the number of labeled objects in object recognition datasets, such as ImageNet~\cite{deng10eccv}.  
Consequently, though modern object recognition systems have the capacity to recognize thousands of object classes, existing state-of-the-art caption models 
lack the ability to form compositional structures which 
integrate new objects with known concepts without explicit examples of image-sentence pairs.
To address this limitation, we propose the Deep Compositional Captioner (\modelName{}) which can 
combine visual groundings of lexical units to
generate  descriptions about objects which are not present in caption corpora (paired image-sentence data), but are present in object recognition datasets (unpaired image data) and text corpora (unpaired text data).

\modelName{} builds on recent deep captioning models \cite{vinyals14arxiv,donahue15cvpr,kirosArxiv14,karpathy14arxiv,mao15iccv} which combine convolutional and recurrent networks for visual description. However unlike previous models which can only describe objects that are present in paired image-sentence data, \modelName{} is compositional in the sense that it can seamlessly construct sentences about new objects by combining them with already seen linguistic expressions in paired training data. 
To illustrate, consider the image of the otter in Figure~\ref{fig:teaser}. To describe the image accurately, any captioning model needs to identify the constituent visual elements such as ``otter'', ``water'' and ``sitting'' and combine them to generate a coherent sentence. While previous deep caption models learn to combine visual elements into a cohesive description exclusively from image and caption pairs, \modelName{} can compose a caption to describe a new visual element such as the ``otter'' 
by understanding that ``otters'' are similar to ``animals'' and can thus be composed in the same way with other lexical expressions. 
To effectively describe new objects, our model incorporates two key design elements.
First, \modelName{} consists of a separate lexical classifier and language model, which can each be trained independently on unpaired image data and unpaired text data. Additionally, the lexical classifier and language model can be combined into a deep caption model which is trained jointly on paired image-sentence data.
Second, and crucial for generating compositional captions,
is the multimodal layer where knowledge from known objects in paired image-sentence datasets can be transferred to new objects only seen in unpaired datasets.
In this work, we leverage external text corpora to relate novel objects to concepts seen in paired data and propose two mechanisms to transfer knowledge from known objects to novel objects.
We demonstrate the ability of \modelName{} to generate captions about new objects by empirically  studying results on a training split of the MSCOCO dataset which excludes certain objects.  Qualitatively, we show that our model can be used to describe a variety of objects in the Imagenet 7k dataset which are not present in caption datasets.  Furthermore, we demonstrate that the efficacy of  \modelName{} is not limited to images, but can also be used to describe new objects in videos by presenting results on a collection of Youtube video clips.

\section{Related Work}

\myparagraph{Deep Captioning}  In the last year, a variety of models \cite{donahue15cvpr,vinyals14arxiv, karpathy14arxiv, kirosArxiv14,fang15cvpr,mao15iccv} have achieved promising results on the image captioning task.
Some \cite{donahue15cvpr,vinyals14arxiv, karpathy14arxiv} follow a CNN-RNN framework: first
high-level features are extracted from a CNN trained on the image classification task, and then a recurrent model learns to predict subsequent words of a caption conditioned on image features and previously predicted words.
Others \cite{kirosArxiv14,mao15iccv} adopt a multimodal framework in which recurrent language features and image features are embedded in a multimodal space.
The multimodal embedding is then used to predict the caption word by word.
Retrieval methods \cite{devlin2015language}
based on comparing the k-nearest neighbors of training and test images in a deep image feature space, have also achieved competitive results on the captioning task.
However, retrieval methods are limited to words and descriptions which appear in a training set of paired image-sentence data.
As opposed to using high level image features extracted from a CNN, another approach  \cite{fang15cvpr,wu2015image} is to train classifiers on visual concepts such as objects, attributes and scenes. A language model, such as an LSTM \cite{wu2015image} or maximum entropy model \cite{fang15cvpr}, then generates a visual description conditioned on the presence of classified visual elements.
Our model most closely resembles the framework suggested in \cite{mao15iccv} which uses a multimodal space to combine features from image and language, however our approach modifies this framework considerably to describe concepts that are never seen in paired image-sentence data.

\myparagraph{Zero-Shot Learning} %
Zero-shot learning has received substantial attention in computer vision \cite{rohrbach10cvpr,ParikhICCV2011,LampertPAMI2014,SocherNIPS2013,frome2013devise} 
since it becomes difficult to obtain sufficient labeled images as the number of object categories grows.
In particular, our method draws on previous zero-shot learning work that mines object relationships from external text data~\cite{rohrbach10cvpr,SocherNIPS2013,frome2013devise}. 
\cite{rohrbach10cvpr} uses text corpora to determine how objects are related to each other, %
then classifies unknown objects based on their relationship to known objects.
In~\cite{SocherNIPS2013,frome2013devise}, images are mapped to semantic word vectors corresponding to their classes, and the resulting image embeddings are used to detect and distinguish between unseen and seen classes.
We also exploit transfer learning via an intermediate-level semantic word vector representation,
however, the above approaches focus specifically on assigning a category label, while our method generates full sentence descriptions. 
In \cite{hoffman14nips}, zero-shot object detectors are learned by transferring information about how network weights trained on the classification task differ from weights trained on the detection task.
We explore a similar transfer method to transfer information from weights which are trained on image-sentence data to weights which are only trained on text data.

\myparagraph{Describing New Objects in Context} 
Many early caption models \cite{thomason14coling,krishnamoorthy13aaai, guadarrama13iccv,kulkarni2013babytalk,gupta09cvpr} rely on first discerning visual elements from an image, such as subjects, objects, scenes, and actions, then filling in a sentence template to create a coherent visual description.
These models are capable of describing objects without being provided with paired image-sentence examples containing the objects, but are restricted to generating descriptions using a fixed, predetermined template.  More recently, \cite{mao15iccv} explore describing new objects with a deep caption model with only a few paired image-sentence examples during training.
However, \cite{mao15iccv} do not consider how to describe objects when no paired image-sentence data is available.
Our model provides a mechanism to include information from existing vision datasets as well as unpaired text data, whereas \cite{mao15iccv} relies on additional image-sentence annotations to describe novel concepts.

\section{Deep Compositional Captioner}
\modelName{} composes novel sentences about objects unseen in paired image-sentence data.
Although it is common to pre-train deep caption models on unpaired image data, unlike existing models, we are able to describe objects present in unpaired image data but not present in paired image-sentence data.
Additionally, to enhance the language structure, we train our model on independent text corpora.
Further, we explore methods to transfer knowledge between semantically related words to compose descriptions of new objects.
Our method consists of three stages: 1) training a deep lexical classifier and deep language model with unpaired data, then, 2) combining the lexical classifier and language model into a caption model which is trained on paired image-sentence data, and, finally, 3) transferring knowledge from words which appear in paired image-sentence data to words which do not appear in paired image-sentence data. 

\subsection{Deep Lexical Classifier}
\label{subsec:method_lexical_class}

The lexical classifier (Fig~\ref{fig:mrnn-diagram}, left) is a CNN which maps images to semantic concepts. %
In order to train the lexical classifier, we first mine concepts which are common in  paired image-text data by extracting the part-of-speech of each word \cite{toutanova2003feature} and then select the most common adjectives, verbs, and nouns.
We do not refine the mined concepts, which means some of the concepts, such as ``use'', are not strictly visual.
In addition to concepts common in paired image-sentence data, the classifier is also trained on 
objects that we wish to describe outside of the caption datasets.

The lexical classifier is trained by fine-tuning a CNN which is pre-trained on the training split of the ILSVRC-2012 \cite{russakovsky2014imagenet} dataset.
When describing images, multiple visual concepts from the image influence the description.
For example, the sentence ``An alpaca stands in the green grass.'' includes the visual concepts ``alpaca'', ``stands'', ``green'', and ``grass''.
In order to apply multiple labels to each image, we use a sigmoid cross-entropy loss.
We denote the image feature output by the lexical classifier as  $f_I$, where each index of $f_I$ corresponds to the probability that a particular concept is present in the image.
Our idea of learning visual classifiers from  text descriptions for captioning is similar to
\cite{rohrbach15gcpr} who learn  classifiers for objects, verbs, and locations and \cite{fang15cvpr} who learn visual concepts using multiple instance learning.

\subsection{Language Model}

\begin{figure}[t]
\begin{center}
\includegraphics[width=1\linewidth]{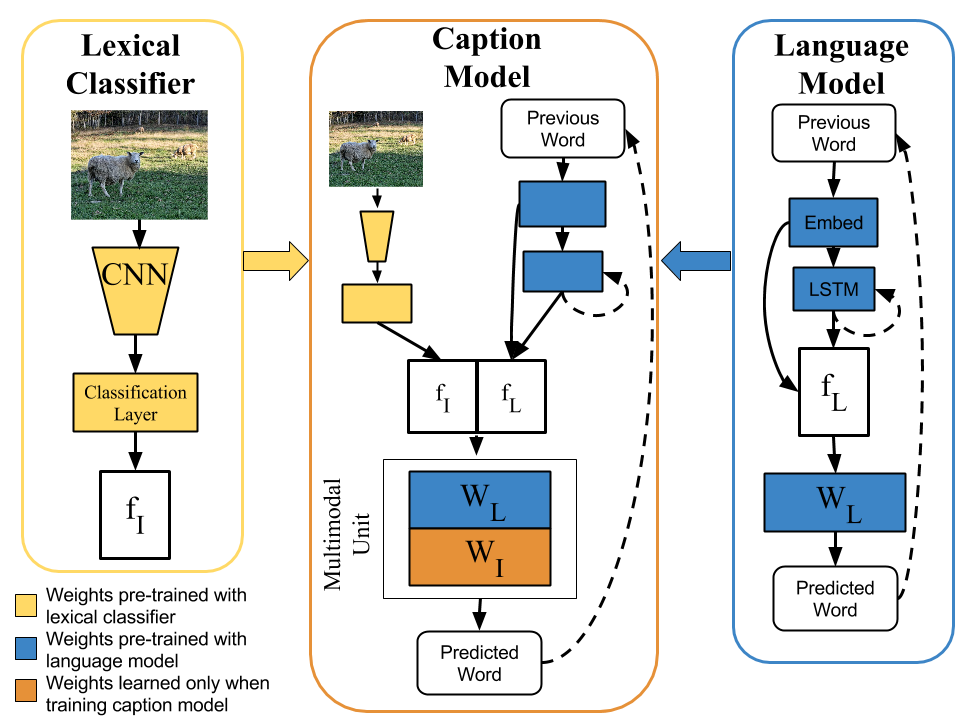}
\end{center}
   \caption{\modelName{} consists of a lexical classifier, which maps pixels to semantic concepts and is trained only on unpaired image data, and a language model, which learns the structure of natural language and is trained on unpaired text data.  The multimodal unit of \modelName{} integrates the lexical classifier and language model and is trained on paired image-sentence data. }
\label{fig:mrnn-diagram}
\end{figure}

The language model (Fig~\ref{fig:mrnn-diagram}, right) learns sentence structure using only unpaired text data and includes an embedding layer which maps a one-hot-vector word representation to a lower dimensional space, an LSTM \cite{hochreiter1997long}, and a word prediction layer.
The language model is trained to predict a word given previous words in a sentence.  
At each time step, the previous word is input into the embedding layer.
The embedded word is input into an LSTM, which learns the recurrent structure inherent in language.
The embedded word and LSTM output are concatenated to form the language features, $f_L$.
$f_L$ is input to an inner product layer which outputs the next word in a generated sequence.
At training time, the ground truth word is always used as an input to the language model, but at test time we input the previous word predicted by our model.
We also find that results improve by enforcing a constraint that the model cannot predict the same word twice in a row.
We explore a variety of sources for unpaired text corpora as described in Section~\ref{sec:train-test-splits}.

\subsection{Caption Model}

The caption model integrates the lexical classifier and the language model to learn a joint model for image description.
As shown in Fig~\ref{fig:mrnn-diagram} (center) the multimodal unit in the caption model combines the image features, $f_I$ and the language features, $f_L$.
The multimodal unit we use is an affine transformation of the image and language features: 
\begin{equation}
p_w = \text{softmax}(f_{I} W_{I} + f_{L} W_{L} + b)
\end{equation}
where $W_{I}$, $W_{L}$, and $b$ are learned weight matrices and $p_w$ is a probability distribution over the predicted word.

Intuitively, the weights in $W_{I}$ learn to predict a set of words which are likely to occur in a caption given the visual elements discerned by the lexical classifier.
In contrast, $W_L$ learns the sequential structure of natural language by learning to predict the next word in a sequence given the previous words.
By summing $f_{I} W_{I}$ and $f_{L} W_{L}$, the multimodal unit combines the visual information learned by the lexical classifier with the knowledge of language structure learned by the language model to form a coherent description of an image.

Both the language model and caption model are trained to predict a sequence of words, whereas the lexical classifier is trained to predict a fixed set of candidate visual elements for a given image.
Consequently, the weights $W_L$, which map language features to a predicted word are learned when training the language model, but the weights $W_I$ are not. 
\camera{Weights in $W_L$ are pretrained using unpaired text data before fine-tuning with paired image-sentence data}, $W_I$ are trained purely with image-sentence data.
Though we use a linear multimodal unit, our results are comparable to results achieved by other methods which include a nonlinear layer for word prediction.  For example, on the MSCOCO validation set \cite{donahue15cvpr} achieves a METEOR score of 23.7, and DCC achieves a METEOR of 23.2.

The caption model is designed to enable easy transfer of learned weights from words which appear in the paired image-sentence data to words which do not appear in the image-sentence data.
First, by using a lexical classifier to extract image features, image features have explicit semantic meaning.
Consequently, it is trivial to expand the image feature to include new objects and to adjust weights in the multimodal unit which correspond to specific objects.
Second, by learning language features using unpaired text data, we ensure that the model learns a good embedding for words which are not present in paired image-sentence data.
Finally, by using a single-layer, linear multimodal unit, the dependence between image and language features and predicted words is straightforward to understand and easy to exploit for semantic transfer.

\subsection{Transferring Information Between Objects}
\label{sec:multimodal}

\begin{figure}[t]
\begin{center}
\includegraphics[width=1\linewidth]{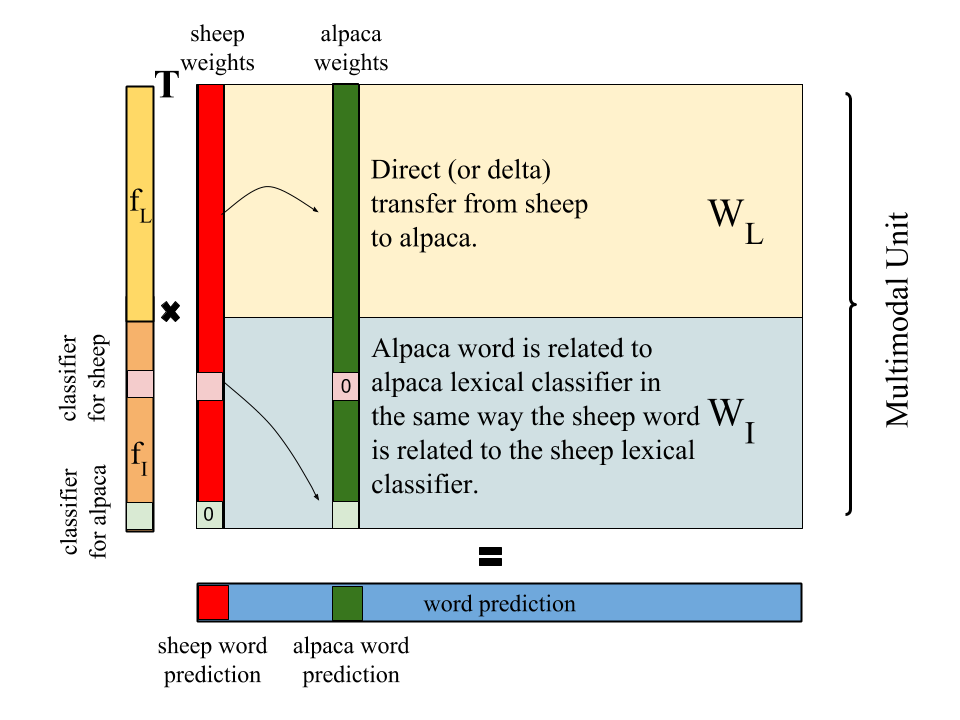}
\end{center}
   \caption{Method for transferring knowledge from words trained with paired image-sentence data to words trained without image-sentence data.  See Section~\ref{sec:multimodal} for details.}
\label{fig:multimodal-unit}
\end{figure}

\myparagraph{Direct Transfer} The first method we explore to transfer weights between objects directly transfers learned weights in $W_{I}$, $W_{L}$ and $b$ from words that appear in the paired image-sentence dataset to words which do not appear in a paired image-sentence dataset (Fig~\ref{fig:multimodal-unit}).
Intuitively, the direct transfer model requires that a new word is described in the same way that semantically similar words are described.
To illustrate, consider the new word ``alpaca'' which is semantically close to the known word ``sheep''.
Let $v_{a}$ and $v_{s}$ indicate the index of the words alpaca and sheep in the vocabulary.
Given image and language features, $f_{I}$ and $f_{L}$ respectively, the probability of predicting the word ``sheep'' is proportional to:
\begin{equation}
f_{I} W_{I}[:,v_{s}] + f_{L} W_{L}[:,v_{s}] + b[v_{s}]
\end{equation}
In order to construct sentences with ``alpaca'' in the same way sentences are constructed with the word ``sheep'', we first directly transfer the weights $W_{I}[:,v_{s}]$, $W_{L}[:,v_{s}]$, and $b[v_{s}]$ (indicated in red in Fig~\ref{fig:multimodal-unit}) to $W_{I}[:,v_{a}]$, $W_{L}[:,v_{a}]$, and $b[v_{a}]$ (indicated in green in Fig~\ref{fig:multimodal-unit}).
Additionally, we expect the prediction of the word ``sheep'' to be highly dependent on the likelihood that a ``sheep'' is present in the image.
In other words, we expect $W_{I}[:,c_{s}]$ to strongly weight the output of the lexical classifier which corresponds to the word ``sheep''.  
However, $W_{I}[:,c_{a}]$ should strongly weight the lexical classifier which corresponds to the word ``alpaca''.  
To enforce this, we set $W_{I}[r_{a},c_{a}] = W_{I}[r_{s},c_{s}]$ where $r_{a}$ and $r_{s}$ indicate the index in the image features which correspond to the alpaca and sheep classifiers respectively.
Finally, we do not expect the output of the word ``alpaca'' to depend on the presence of a sheep in the image and vice versa.
Consequently, we set $W_{I}[r_{s},c_{a}] = W_{I}[r_{a},c_{s}] = 0$.
\newline

\myparagraph{Delta Transfer}
Instead of directly transferring weights, we can also transfer \textit{how weights change} when trained on paired image-text data.
Again, consider transferring the word ``sheep'' to the word ``alpaca''.
We determine $\Delta_L$ for  a given word as:
\begin{equation}
\Delta_{L} = W_{L-caption}[:,v_{s}] - W_{L-language}[:,v_{s}]
\end{equation}
where $W_{L-caption}$ are weights learned when training with both images and sentences and $W_{L-language}$ are weights learned when training only with language.
The weights for the new word ``alpaca'' are updated as:
\begin{equation}
W_{L-caption}[:,v_{a}] = W_{L-language}[:,v_{a}] + \Delta_{L}
\end{equation}
\camera{
Delta transfer may be advantageous because, unlike direct transfer, it does not overwrite pretrained weights in $W_L$ during transfer.}
When performing delta transfer for $W_L$, we still use direct transfer for weights in $W_I$.

\myparagraph{Determining Concept Similarity}
\label{sec:concept-similarity}
Determining which words in the paired image-sentence data are semantically similar to words out of the paired image-sentence data is key for transfer.
We determine semantic similarity with the word2vec~\cite{Mikolov13} CBOW model which we trained on the British
National Corpus (BNC), UkWaC, and Wikipedia, and estimate word similarity using cosine distance.
Additionally, we restrict words that are transferred to new words to be in the lexical layer.

\section{Experimental Framework}

\subsection{Datasets}
\label{sec:train-test-splits}
\myparagraph{Image Description} To empirically evaluate our method we create a subset of the MSCOCO~\cite{coco2014} training set (denoted as the held-out MSCOCO training set) which excludes all image-sentence pairs which describe at least one of eight MSCOCO objects.  
To ensure that excluded objects are at least similar to some included ones, we cluster the 80 objects annotated in the MSCOCO segmentation challenge using the vectors from the word2vec embedding described in Section~\ref{sec:concept-similarity} and exclude one object from each cluster.
The following words are chosen:  ``bottle'', ``bus'', ``couch'', ``microwave'', ``pizza'', ``racket'', ``suitcase'', and ``zebra''.  
We randomly select 50\% of the MSCOCO validation set for validation, and set aside the other 50\% for testing. 
We use the validation set to determine all model hyperparameters, and present all results on the test set.
We label the visual concepts in each image based on the five ground truth caption annotations provided in the MSCOCO dataset.
If any of the ground truth captions mention an object, the corresponding image is considered a positive example for that object.

In addition to empirically evaluating our model, we also qualitatively examine the performance of \modelName{} at a large scale by describing objects outside of the paired image-sentence corpora.
Specifically, we select 642 objects from the full ImageNet object recognition dataset \cite{deng10eccv} which do not occur in MSCOCO and are also present in the WebCorpus text dataset (see section~\ref{sec:language-model-training}) vocabulary.
We do no manual concept pruning; consequently some selected concepts refer to a broad variety of objects (e.g., the class ``fauna'' contains all animals) and other classes only contain a small number of images (e.g., there are three ``discus'' images).
We use %
75\% of images from each class to train the lexical classifier, and evaluate on the rest. 
We stress that we do not have any descriptions for these categories.

\myparagraph{Video Description} For empirical evaluation on video description, we use a collection of Youtube clips from the Microsoft Video description (MSVD) corpus \cite{chen:acl11}, which contains 1,970 short annotated clips. 
Our basic experimental setting %
follows previous video description works \cite{venugopalan15naacl,S2VT}. However, %
we hold out paired video-sentence data for some objects during training.
Because there is significant variation in the number of video clips containing each object in the MSVD dataset, 
we hold out objects in the MSVD dataset which appear in five or fewer training videos and at least one test video and also appear in the ILSVRC2015 video object detection challenge set.\footnote{\url{http://image-net.org/challenges/LSVRC/2015/}}
Our MSVD held-out set excludes paired video-sentence training data which include ``zebra'', ``hamster'', ``broccoli'', and ``turtle''.

We also qualitatively evaluate our method on the ILSVRC object detection challenge  videos (initial release) which consists of 1,952 video snippets of the 30 objects from the ILSVRC2015 object detection in video. 
Objects which we describe in the detection challenge videos include ``whale'', ``fox'', ``hamster'', ``lion'', ``zebra'', and ``turtle''.

\subsection{Training the Lexical Classifier}

\myparagraph{Image description} We consider both MSCOCO and ImageNet as sources of labeled image data to train the lexical classifier.
For all objects in paired image-sentence data, we use COCO images which are labeled with 471 visual concepts to train the lexical classifier.
For the eight objects which do not appear in the paired image-sentence data, we explore training the lexical classifier using MSCOCO images (in-domain) and ImageNet images (out-of-domain).
For qualitative experiments on ImageNet objects, we use Imagenet images to train the lexical classifier on new visual concepts.
The lexical classifier is trained by fine-tuning a deep convolutional model (VGG-16 layer \cite{vgg}) trained on the ILSVRC-2012 \cite{russakovsky2014imagenet} object recognition training subset of the ImageNet dataset. 

\myparagraph{Video Description} Unlike images, videos consist of a sequence of frames which need to be mapped to a set of semantic concepts by the lexical classifier.
To build a lexical classifier for videos, we mean-pool $fc_7$ features across all frames in a video clip before classification.
We use both MSVD and ImageNet videos to train the lexical classifier.  
We use the VGG-16 layer model to extract $fc_7$ layer features from video frames. 

\subsection{Training the Language Model}
\label{sec:language-model-training}
\myparagraph{Image Description} We consider three different sources for unpaired text data to train the language model:
(1)  \textbf{MSCOCO} consists of all captions from the MSCOCO train set %
(2)  \textbf{Text from Image Description Corpora (CaptionTxt)} consists of text data from other paired image and video description datasets: Flickr1M~\cite{huiskes08}, Flickr30k~\cite{flickr30k}, Pascal-1k~\cite{rashtchian2010collecting} and ImageCLEF-2012~\cite{thomee2012overview} and sentence descriptions of Youtube clips from the MSVD training corpus.  
This corpus \textit{does not} include sentences from MSCOCO.
(3) \textbf{External text (WebCorpus)} consists of 60 million sentences from the British National Corpus (BNC), UkWaC, and Wikipedia. 

\myparagraph{Video Description}
We consider two sources of text to train the video description language model.  
The first is the WebCorpus text described above.  
We also consider a slight variant on the CaptionTxt described above which includes descriptions from MSCOCO, Flickr-30k~\cite{flickr30k}, Pascal-1k~\cite{rashtchian2010collecting} and the MSVD sentence descriptions.

\subsection{Training the Caption Model} 
After training the lexical classifier and language model, the weights in the multimodal layer of the caption model are trained with paired image-sentence data.
For the direct transfer method, we simply train the weights in the multimodal unit ($W_I$ and $W_L$) while freezing all other weights.
For the delta transfer method, if weights in $W_L$, which are pretrained when training the language model, diverge too much from their original values, transfer does not work well.  Consequently, we first hold weights in $W_L$ constant, training only $W_I$, before jointly training $W_L$ and $W_I$.  
The caption model is trained the same way for both image and video description.

\subsection{Metrics}
\begin{figure}[t]
\begin{center}
\includegraphics[scale=0.5]{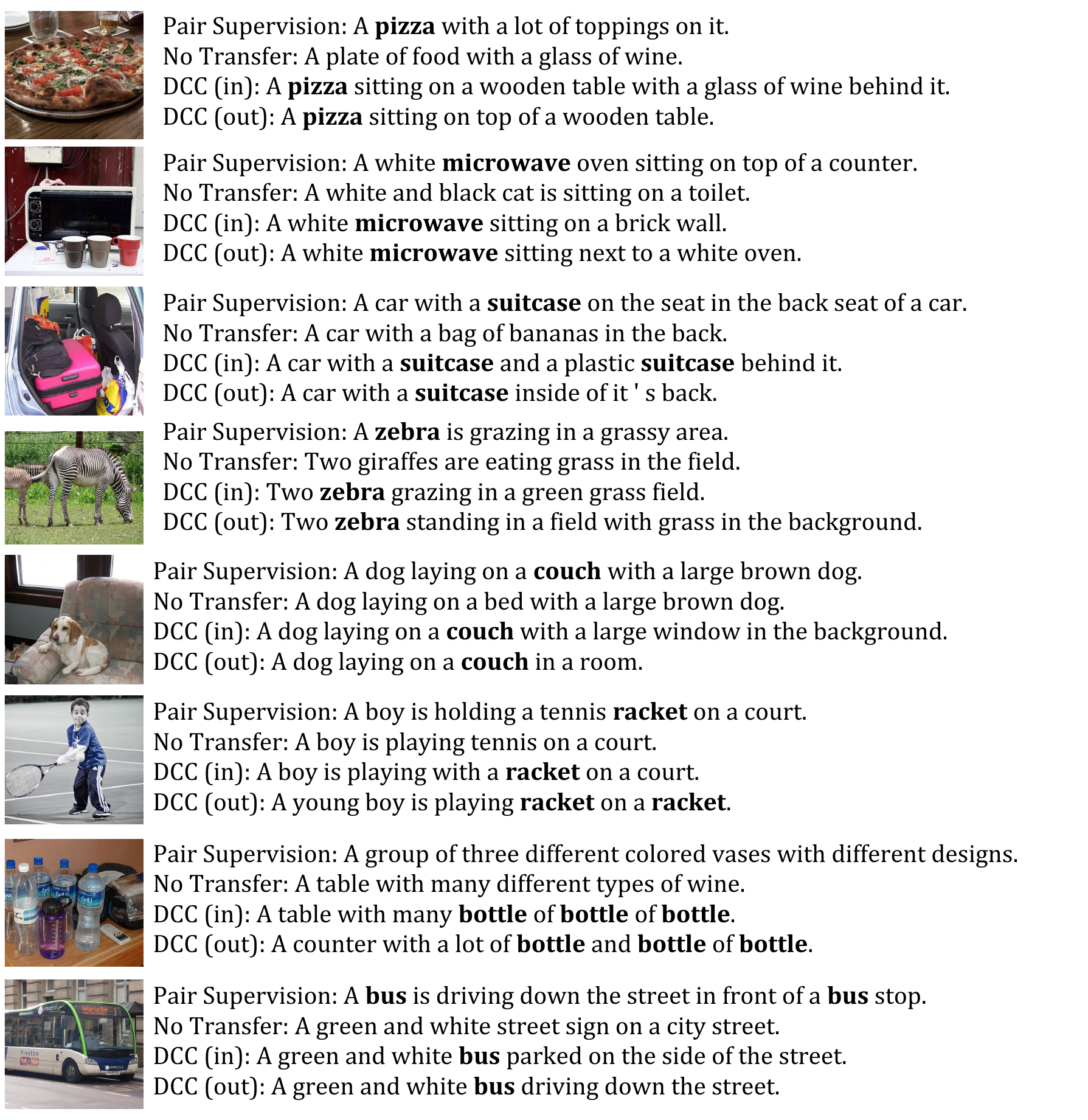}
\end{center}
    \caption{Image Description: Comparison of captions generated by model without transfer, DCC with in-domain training (MSCOCO), with out-of-domain training (ImageNet and WebCorpus), and a model trained with paired image-sentence supervision for all MSCOCO objects.  DCC is capable of integrating new words and generates sentences similar to those generated when paired image-sentences for all objects are present during training.}
    \label{table:qual-MSCOCO}
\end{figure}

To evaluate our transfer methods, we must choose a metric that indicates whether or not a generated sentence includes a new object.  Common caption metrics such as BLEU \cite{papineni2002bleu} and METEOR \cite{banerjee2005meteor} measure overall sentence meaning and fluency.  However, for many objects, it is possible to achieve good BLEU and METEOR scores without mentioning the new object (e.g., consider sentences describing the boy playing tennis in Figure~\ref{table:qual-MSCOCO}).  
To definitively report our model's ability to integrate new vocabulary, we also report the F1-score.
The F1-score considers ``false positives'' (when a word appears in a sentence it should not appear in), ``false negatives'' (when a word does not appear in a sentence it should appear in), and ``true positives'' (when a word appears in a sentence it should appear in).
We consider generated sentences ``positive'' if they contain at least one mention of a held out word and ground truth sentences ``positive'' if a word is mentioned in any ground truth annotation that describes an image.

We train our models using \textit{Caffe}\cite{jia2014caffe}.    \footnote{Code can be found at \url{http://www.eecs.berkeley.edu/~lisa\_anne/dcc\_project\_page.html}.}

\vspace{-0.2cm}
\section{Results}

\label{sec:direct-delta}

\begin{table}
\begin{center}
\begin{tabular}{|l|cccc|}
\hline
 & LRCN~\cite{donahue15cvpr} & No Transfer& $\Delta$T  & DT \\

 \hline
 F1 & 0 & 0 & 34.89  & \textbf{39.78} \\
 BLEU-1 & 63.65 & 62.99 & 64.00 & \textbf{64.40} \\
 METEOR & 19.33 & 19.9 & 20.86 & \textbf{21.00} \\
\hline
\end{tabular}\\
\end{center}
\caption{We compare DCC before transfer (No Transfer) to DCC with delta transfer ($\Delta$T) and DCC with direct transfer (DT).
We also compare to another competitive caption generation model (LRCN).
We measure our models ability to insert new words into a generated sentence with the F1-score.
We also report Bleu-1 and METEOR, which indicates overall sentence quality.  \modelName{} successfully incorporates new words and improves sentence quality.
(Values in \%)}
\label{table:direct-delta-f1}
\end{table}

\begin{table*}
\begin{center}
\begin{tabular}{|c|c|c|c|c|c|c|c|c|c|c|c|c|}
\hline
& bottle & bus & couch & microwave & pizza & racket & suitcase & zebra & average \\ \hline
Pair Supervision & 23.20 & 72.07 & 50.60 & 39.48 & 77.07 & 38.52 & 46.50 & 91.02 & 54.81\\
DT & 4.63 & 29.79 & 45.87 & 28.09 & 64.59 & 52.24 & 13.16 & 79.88 & 39.78 \\
\hline
\end{tabular}\\
\end{center}
\caption{Image Description: Comparison of F1 scores for direct transfer DCC model (DT) and a model trained with image-sentence training examples for all objects. (Values in \%) }
\label{table:F1}
\end{table*}

\begin{table}
\begin{center}
\begin{tabular}{|c|c|ccc|}
\hline
Lexical  & Language & B-1 & METEOR & F1 \\
classifier & model & & & \\
\hline
MSCOCO & MSCOCO & 64.40 & 21.00 & 39.78 \\
Imagenet & MSCOCO &  64.00 & 20.71 & 33.60 \\
Imagenet & CaptionTxt & 64.79 & 20.66 & 35.53 \\
Imagenet & WebCorpus & 64.85 & 20.66 & 34.94 \\
\hline
\end{tabular}\\
\end{center}
\caption{Image Description: We compare the effect of pre-training the lexical classifier and language model with different unpaired image and text data sets.  As expected, we see the best result when using in domain MSCOCO data to train the lexical classifier and language model, though training with out of domain corpora is comparable. (Values in \%)}
\label{table:dst}
\end{table}

\subsection{Image Description}

As shown in Figure~\ref{table:qual-MSCOCO}, \modelName{} is capable of integrating new vocabulary into image descriptions in a cohesive manner.

\myparagraph{Direct Transfer Versus Delta Transfer} Table~\ref{table:direct-delta-f1} compares the average F1-score across the eight held-out training classes for the delta transfer and direct transfer methods.
\camera{We additionally train LRCN (\cite{donahue15cvpr}) \footnote{For fair comparison, we train LRCN on VGG, fine-tune through the entire network, and do not use beam search.} on our MSCOCO held-out dataset and note that our model without transfer yields comparable results to LRCN, and performs considerably better on all metrics after transfer.}
As shown by the F1-scores reported in Table~\ref{table:direct-delta-f1}, both the delta transfer and direct transfer methods are capable of integrating new words into their vocabulary.  
We also report the BLEU-1 score, which measures the overlap between generated words and words in reference sentences.
By measuring the METEOR score, we ensure that our model maintains sentence fluency when inserting new objects.
\modelName{} consistently increases METEOR scores indicating that overall sentence quality improves with \modelName{}.
The direct transfer method improves F1-scores, BLEU, and METEOR scores by a larger amount than the delta transfer method and is thus used for the remainder of our experiments.

Importantly, BLEU and METEOR scores do not decrease for objects which are present in the held-out training data set.
When trained with all image-sentence training examples, our model achieves an average BLEU-1 of 69.36 and METEOR of 23.98 on held-out classes.

To illustrate which words our model works best on, we report the F1-score for individual objects in Table~\ref{table:F1}. 
We compare to a model which is trained with image-sentence pairs for the eight held-out objects.
For all objects, \modelName{} is able to compose sentences which include the object.

\myparagraph{Analysis of Transfer Words} In general, determining word similarity with a word2vec embedding works well.  Words such as ``zebra''/``giraffe'' and ``microwave''/``refrigerator'' are close in embedding space and are also used in similar ways in natural language, suggesting they will work well for transfer.
Some transfer pairs (``racket''/``tennis'' and ``bus''/``stop'') are used together frequently but play different structural roles in sentences.
Consequently, the word ``racket'' is frequently used like the word ``tennis'' leading to grammatical errors.
However, similar errors do not occur when transferring ``stop'' to ``bus''.

\myparagraph{Pre-Training with Out-of-Domain Data} In the above experiments the lexical classifier and language model are pre-trained using MSCOCO images and text.
In a real world scenario, it is unlikely that available unpaired image and text data will be from the same domain as paired image-sentence data.
However, it is essential that the model learns good image and language features.
Naturally, if the lexical classifier is unable to recognize certain objects, \modelName{} will not be able to describe the objects.
Perhaps more subtly, if the language model is not trained with unpaired text which includes an object, it will not learn a proper embedding for the new word and will not produce cohesive descriptions about new objects.

Table \ref{table:dst} demonstrates the impact of using outside image and text corpora to train the lexical classifier and language model.
Our model performs best when provided with in-domain image and text for all training stages, but performance is comparable when using ImageNet images to train the lexical classifier and CaptionTxt or WebCorpus text data to train the language model.

\subsection{Describing ImageNet Objects}
\label{sec:qual-imagnet}

\begin{figure}[]
\begin{center}
\includegraphics[scale=0.5]{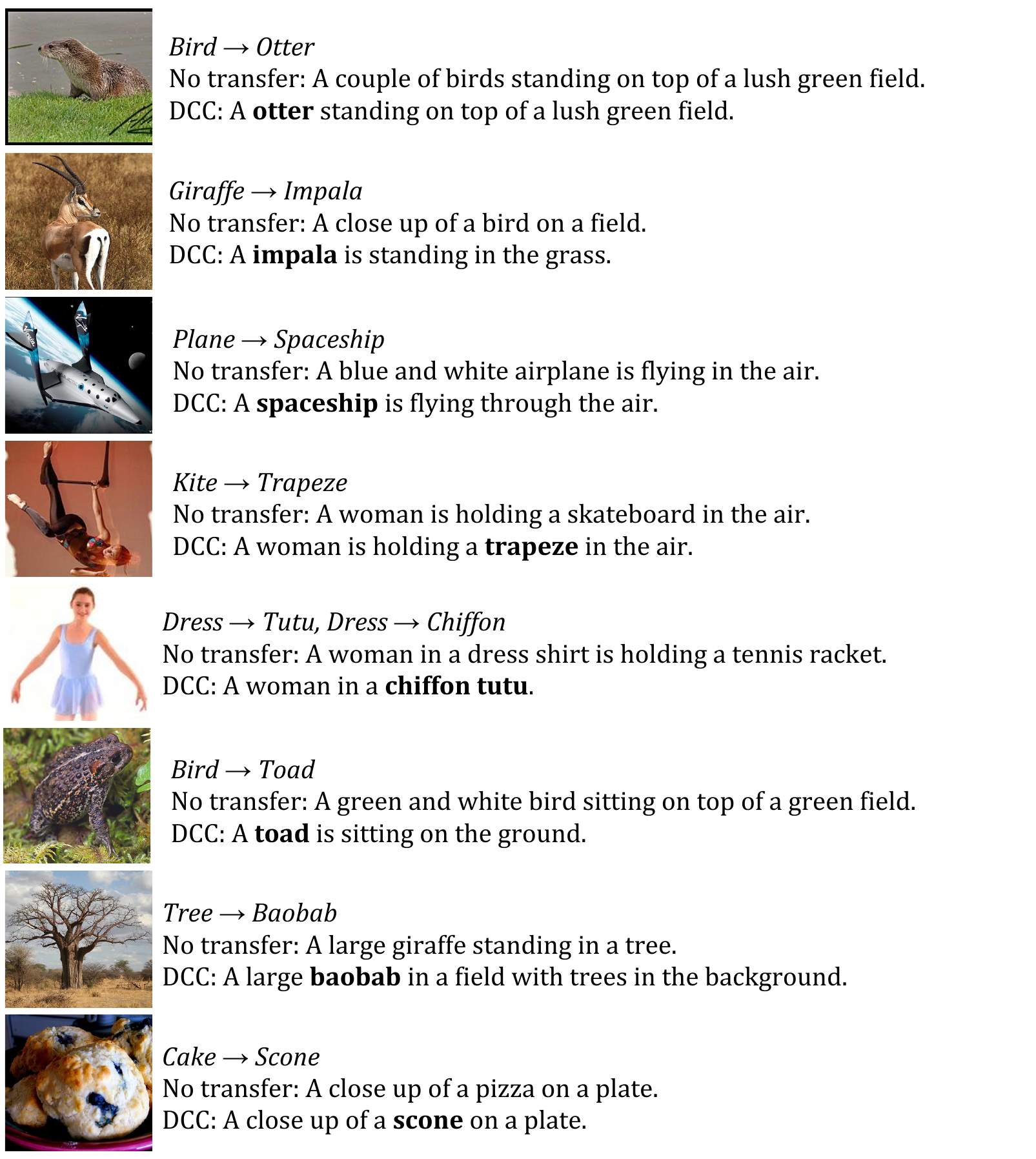}
\end{center}
    \caption{Image Description: DCC is able to describe Imagenet objects (bolded) which are not mentioned in any of the paired image-sentence data, and therefore cannot be described by existing deep caption models.  X $\rightarrow$ Y indicates that the known word X is transferred to the new word Y.}
    \label{tbl:res:qual}
\end{figure}

We qualitatively assess our model by describing a variety of ImageNet objects which are not included in the MSCOCO data set (Fig~\ref{tbl:res:qual}).
\modelName{} accurately describes \camera{335 new words including entry-level words like ``toad'' as well as fine-grained categories like ``baobab''.  Though most Imagenet words we transfer are nouns, we are able to successfully transfer some adjectives such as ``chiffon''.  
\modelName{} achieves more than simple noun replacement.  
For example, the sentence ``A large giraffe standing in a tree'' changes significantly to ``A large baobab in a field with trees in the background'' after transfer.
Importantly, our model is able to compose sentences by placing objects in the correct context.  
For example, comparing Fig~\ref{tbl:res:qual} (top) to the image in Fig~\ref{fig:teaser}, the object ``otter'' is correctly described as either ``sitting in the water'' or ``standing on top of a lush green field'' depending on visual context.}

\begin{figure}[t]
\begin{center}
\includegraphics[scale=0.59]{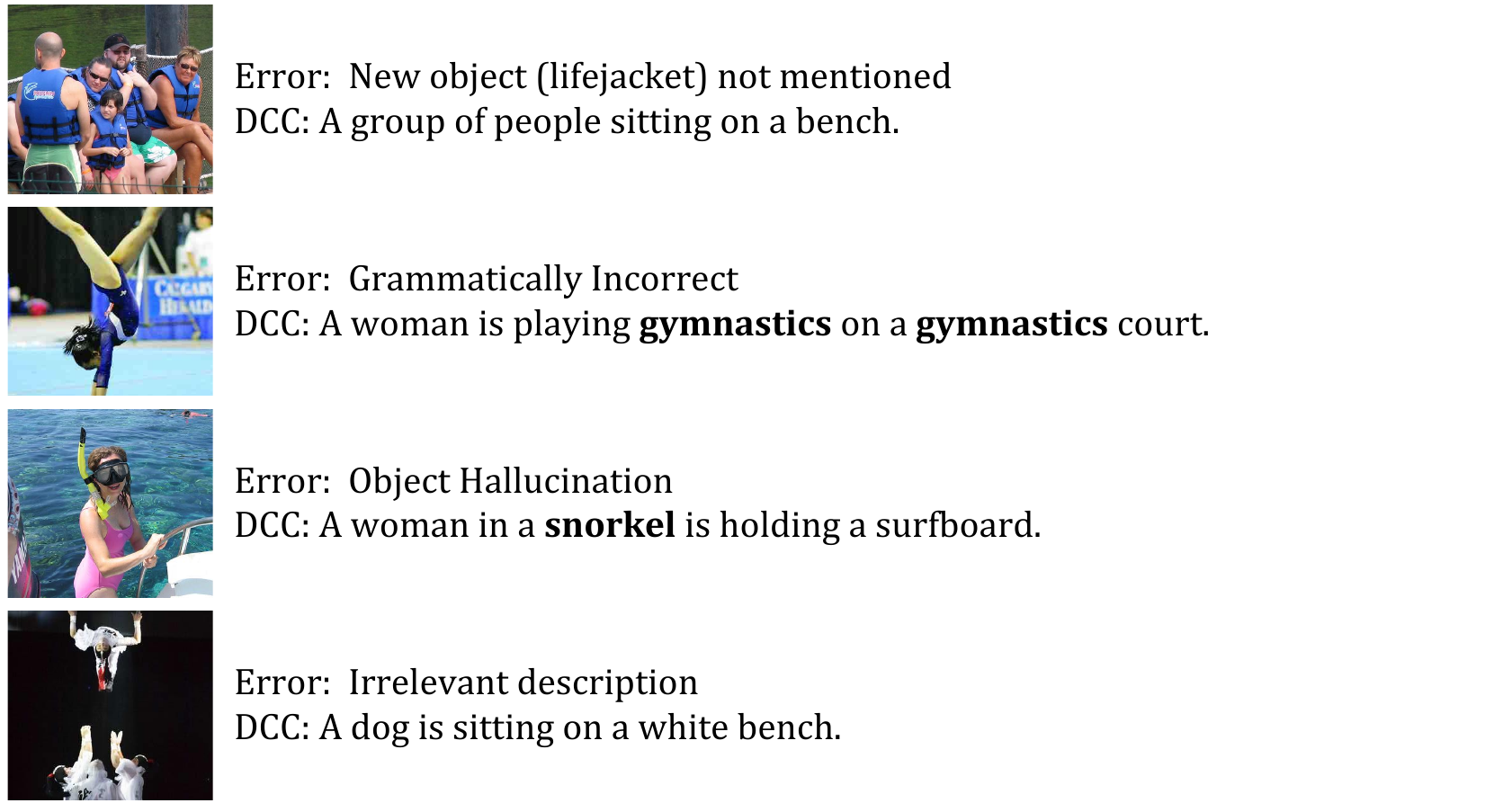}
\end{center}
    \caption{Image Description: We highlight four  common error types generated by the \modelName{}.  See Section~\ref{sec:qual-imagnet} for details.}
    \label{tbl:res:error}
\end{figure}

Figure~\ref{tbl:res:error} examines a few common error types:

\myparagraph{New Object Not Mentioned} (Figure~\ref{tbl:res:error}, top) For some images, \modelName{} produces relevant sentences, but fails to mention the new object.

\myparagraph{Grammatically Incorrect} (Figure~\ref{tbl:res:error}, second row)
Some sentences incorporate new words, but are grammatically incorrect.  For example, 
though \modelName{} describes sentences with the word ``gymnastics'', the resulting sentences are frequently grammatically incorrect (e.g., ``A woman playing gymnastics on a gymnastics court'').  This is likely because the word ``tennis'' is transferred to ``gymnastics''.  Though both of these words are sports, one does not ``play'' gymnastics and gymnastics is not performed on a ``court.''

\myparagraph{Object Hallucination} (Figure~\ref{tbl:res:error}, third row)
\modelName{} frequently hallucinates objects which commonly occur in a specific visual context.  For example, in a beach image, the model commonly includes the word ``surfboard''.  

\myparagraph{Irrelevant Description} (Figure~\ref{tbl:res:error}, bottom)
Some captions do not mention any salient objects correctly.  Such errors can be caused by poor image recognition or because the language model is unable to construct a reasonable sentence from constituent visual elements.

More examples are in our supplemental material.

\subsection{Video description}
\begin{figure}[]
\begin{center}
\includegraphics[scale=0.7]{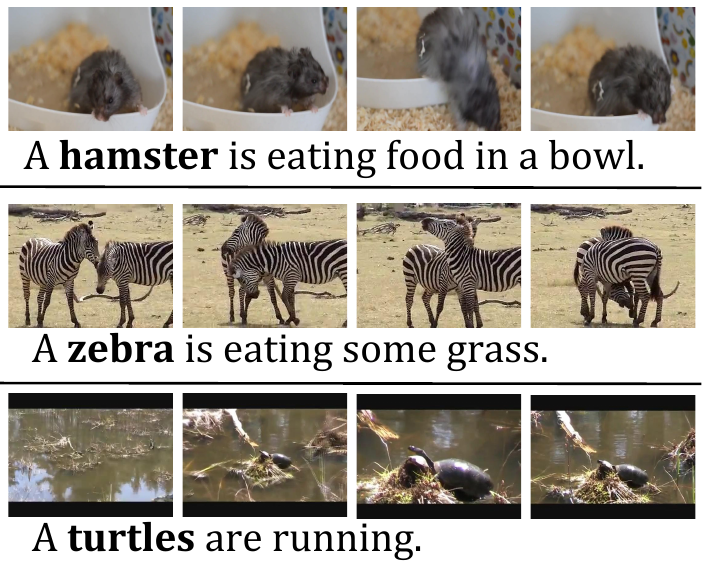}
\end{center}
    \caption{Video Description: Captions generated by DCC on videos of novel objects unseen in paired training data.}
    \label{fig:qual-Video}
\end{figure}

\camera{We believe \modelName{} can be especially beneficial in domains, such as video description, where the amount of paired training data is small.}
Table \ref{tab:yt} presents empirical results of direct transfer \modelName{} on videos in the MSVD corpus (Section~\ref{sec:train-test-splits}). We report the average F1 score on all held-out classes, and METEOR scores on the complete test dataset. As seen by the F1 score, transferring weights allows us to describe new objects in video.  Additionally, the METEOR score improves with transfer demonstrating that \modelName{} improves overall sentence quality. Similar to the trend seen for image captioning, %
training on in-domain text corpora achieves slightly better performance than training on external text. When adding 
ImageNet videos, both F1 and METEOR increase suggesting that including outside image data is beneficial.  
Including ImageNet videos to learn better lexical classifiers especially improves the F1 score, which increases from 6.0 to 22.2.
Figure \ref{fig:qual-Video} presents qualitative results of our best model on snippets with the held out objects in MSVD corpus and the ILSVRC validation set.

\begin{table}
\begin{center}
\begin{tabular}{|l|c|c|}
\hline
Model (Video) & METEOR & F1 \\ \hline
Baseline (No Transfer) &  28.8 & 0.0 \\ 
 + DT &  28.9 & 6.0\\
 + ILSVRC Videos (No Transfer) &  29.0 & 0.0\\ 
 + ILSVRC Videos + DT &  29.1 & 22.2 \\ \hline
\end{tabular}
\end{center}
\caption{Video Description: METEOR scores across the test dataset and average F1 scores for the four held-out categories (All values in \%) using direct transfer (DT). The DCC models were trained on videos with 4 objects removed and the language model was trained on in-domain sentences.}
\label{tab:yt}
\end{table}

\section{Conclusion}
We present the Deep Compositional Captioner (\modelName{}) which can be used to describe new objects which are not present in current caption copora.
Our quantitative and qualitative results demonstrate our model's ability to integrate new vocabulary into generated image and video descriptions \camera{by effectively using existing vision datasets and unpaired text data}.
By integrating data from disparate sources and transferring knowledge between semantically related concepts, \modelName{} improves upon current deep caption models by providing rich descriptions which are not limited by the availability of paired image-sentence corpora.

\section*{Acknowledgements}
Lisa Anne Hendricks is supported by the NDSEG Fellowship.
Marcus Rohrbach was supported by a fellowship within the FITweltweit-Program
of the German Academic Exchange Service (DAAD).
Trevor Darrell was supported in part by DARPA; AFRL; DoD MURI award N000141110688; NSF awards  IIS-1212798, IIS-1427425, and IIS-1536003, and the Berkeley Vision and Learning Center.
Raymond Mooney and Kate Saenko were supported in part by DARPA under AFRL grant FA8750-13-2-0026 and a Google Grant. Mooney was also supported by ONR ATL Grant N00014-11-1-010.

\appendix
\section*{SUPPLEMENTAL MATERIAL}
We present further empirical and qualitative results for both image and video description.  
For the image description task, we explore averaging weight vectors before transfer, illustrate errors made by the model when no unpaired text data is used during training and provide descriptions generated by \modelName{} for a large variety of novel object categories in ImageNet.
For the video description task, we include results when training the language model with an external text corpora and more \modelName{} descriptions of novel objects in video.

\section{Image Description}

\subsection{Transferring from Multiple Words}

In the main paper, we transfer the most similar word in the paired image-sentence data to new objects in the multimodal unit. 
However, we could also average weights across multiple similar words before transfer.
For direct transfer, averaging weights before transfer hurts performance substantially.
In contrast, averaging weights before delta transfer does not significantly impact results (Table ~\ref{tab:delta}).

\subsection{Transfer with No Language model}
\begin{figure}[t]
\begin{center}
\includegraphics[scale=0.6]{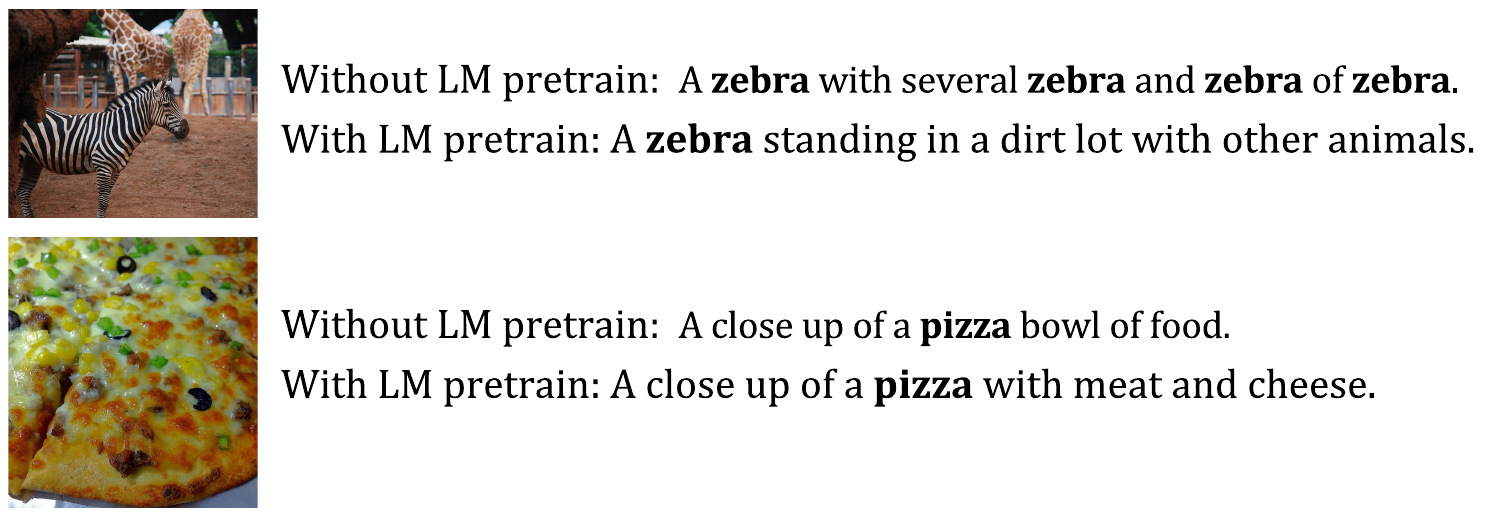}
\end{center}
    \caption{Image Description: Example captions generated by DCC direct transfer with and without pretraining the language model with unpaired text data.  Without pretraining on text data, generated sentence fluency is poor.  For example, the model will repeat the word ``zebra'' or insert a ungrammatical phrase like ``pizza bowl of food''.}
    \label{fig:error_no_lm}
\end{figure}

As mentioned in the paper, if unpaired text data is not used to train the language model, descriptions are poor because the caption model never learns good language features for new object categories. 
This is illustrated in Figure~\ref{fig:error_no_lm}.  
Though a model which is trained on paired image-sentence data and unpaired image data can insert new words into a sentence, the generated sentences are not cohesive because the underlying language model has never seen the new object categories.  
For example, without training on unpaired text data, the model produces repetitive sentences like ``A zebra with several zebra and zebra of zebra.'' or ungrammatical phrases like ``a pizza bowl of food''.

\subsection{Qualitative Analysis of ImageNet Descriptions}

\begin{figure*}
\begin{center}
\includegraphics[scale=0.55]{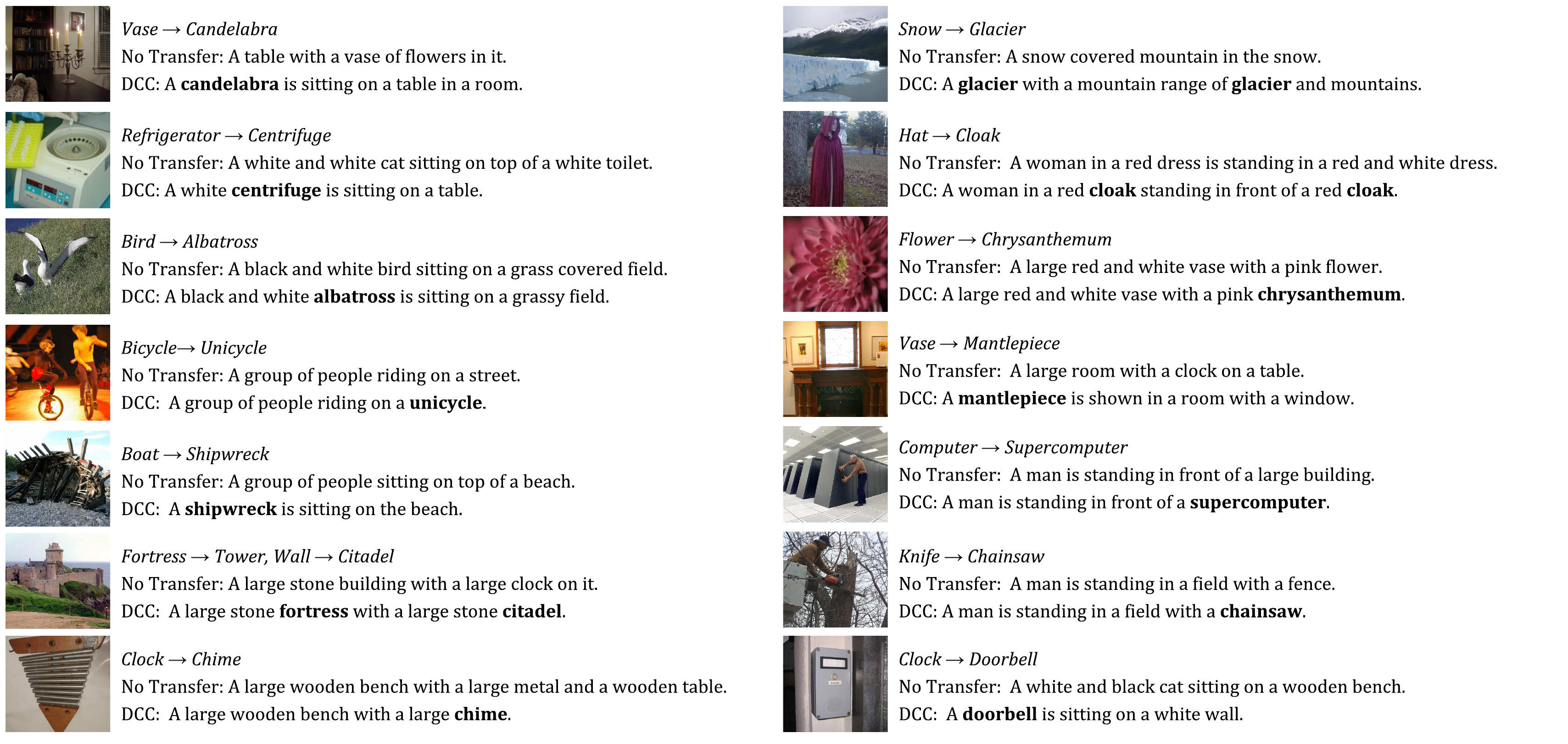}
\end{center}
    \caption{Image Description: Comparison of descriptions generated by \modelName{} for a variety of ImageNet objects with and without transfer.  X $\rightarrow$ Y indicates that the known word X is transferred to the new word Y.  \modelName{} does not simply substitute new words in place of words present in image-text data.  Further, the sentences generated by a model without transfer do not need to contain the word X for the sentences generated by a model after transfer to include the word Y.}
    \label{fig:image_comparison_descriptions}
\end{figure*}

\paragraph{Comparison to sentences generated with no transfer}
Figure~\ref{fig:image_comparison_descriptions} compares generated sentences for new object categories before and after transfer and also indicates which word in the paired image-sentence data is transferred to each new word.  
\modelName{} does not simply substitute words seen in the image-sentence data with new object categories.
Rather, subsequent words in the sentence are impacted by the use of a new vocabulary word.
Consider the image of the candelabra in the top row of Figure~\ref{fig:image_comparison_descriptions}. 
Without transfer, the model describes ``A table with a vase of flowers in it.''.  
However, after transfer, the model describes ``A candelabra is sitting on a table in a room.''
Though the word vase is transferred to the word candelabra, candelabra is described differently.
This is possibly because the language model learns to describe the words ``vase'' and ``candelabra'' in slightly different ways.
For example, ``flowers'' are not frequently in candelabras.
Furthermore, sentences generated before transfer do not necessarily include the known word which is transferred to the new category.
Consider the image of the centrifuge in the second row of Figure~\ref{fig:image_comparison_descriptions}.
Though the word ``refrigerator'' is transferred to the word ``centrifuge'', the sentence generated before transfer does not include the word ``refrigerator''. However, the word ``centrifuge'' is accurately described after transfer.
\begin{table}[t]
\begin{center}
\begin{tabular}{|l|ccc|}
\hline
 & $\Delta$T (N=1) & $\Delta$T  (N=5) & $\Delta$T  (N=15) \\
 
 \hline
 F1 & 34.89 & 34.60 & 34.98 \\
 BLEU-1 & 64.00 & 63.96 & 63.95\\
 METEOR  & 20.86 & 20.88 & 20.88\\
\hline
\end{tabular}
\end{center}
\caption{Image Description: Comparison of delta transfer method when averaging $N$ closest weight vectors before transfer.  Averaging weight vectors before transfer has little impact on performance.}
\label{tab:delta}
\end{table}

\begin{figure*}
\begin{center}
\includegraphics[scale=0.65]{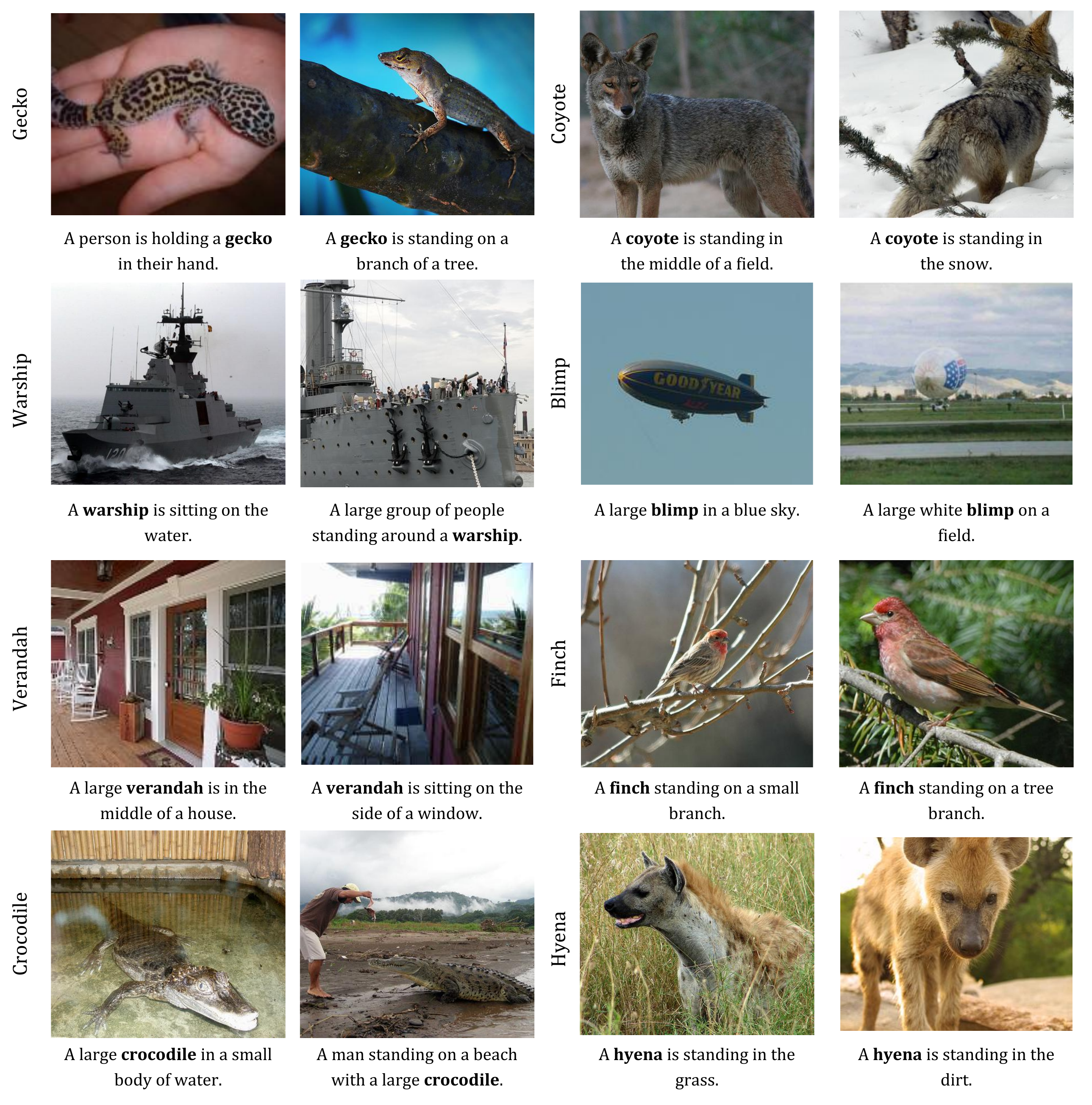}
\end{center}
    \caption{Image Description: Example \modelName{} descriptions for eight different objects.  \modelName{} is able to describe objects in different contexts.}
    \label{fig:many_descriptions1}
\end{figure*}

\begin{figure*}
\begin{center}
\includegraphics[scale=0.65]{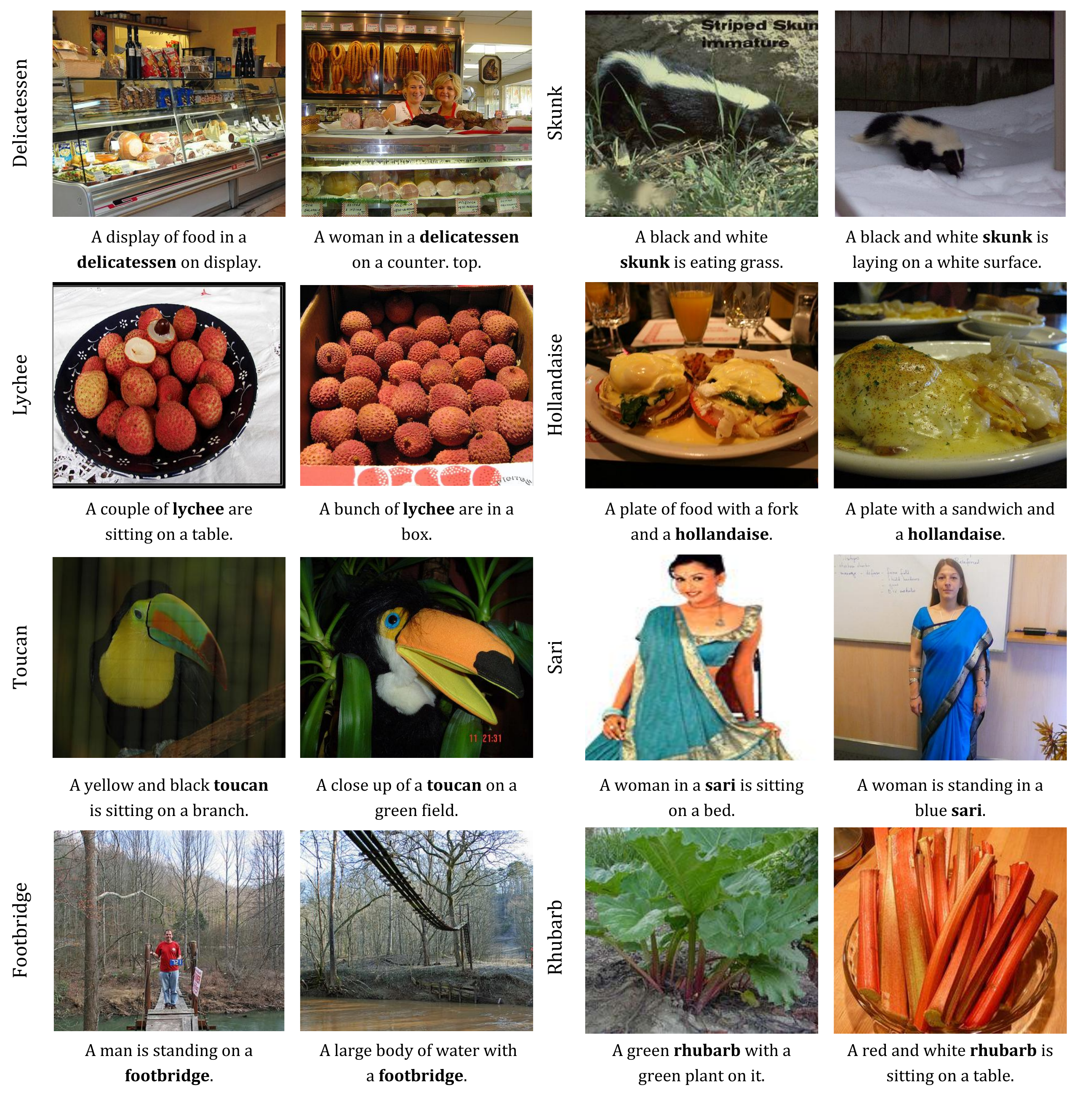}
\end{center}
    \caption{Image Description: Example \modelName{} descriptions for eight different objects.  \modelName{} is able to describe objects in different contexts.}
    \label{fig:many_descriptions2}
\end{figure*}

\begin{figure*}
\begin{center}
\includegraphics[scale=0.65]{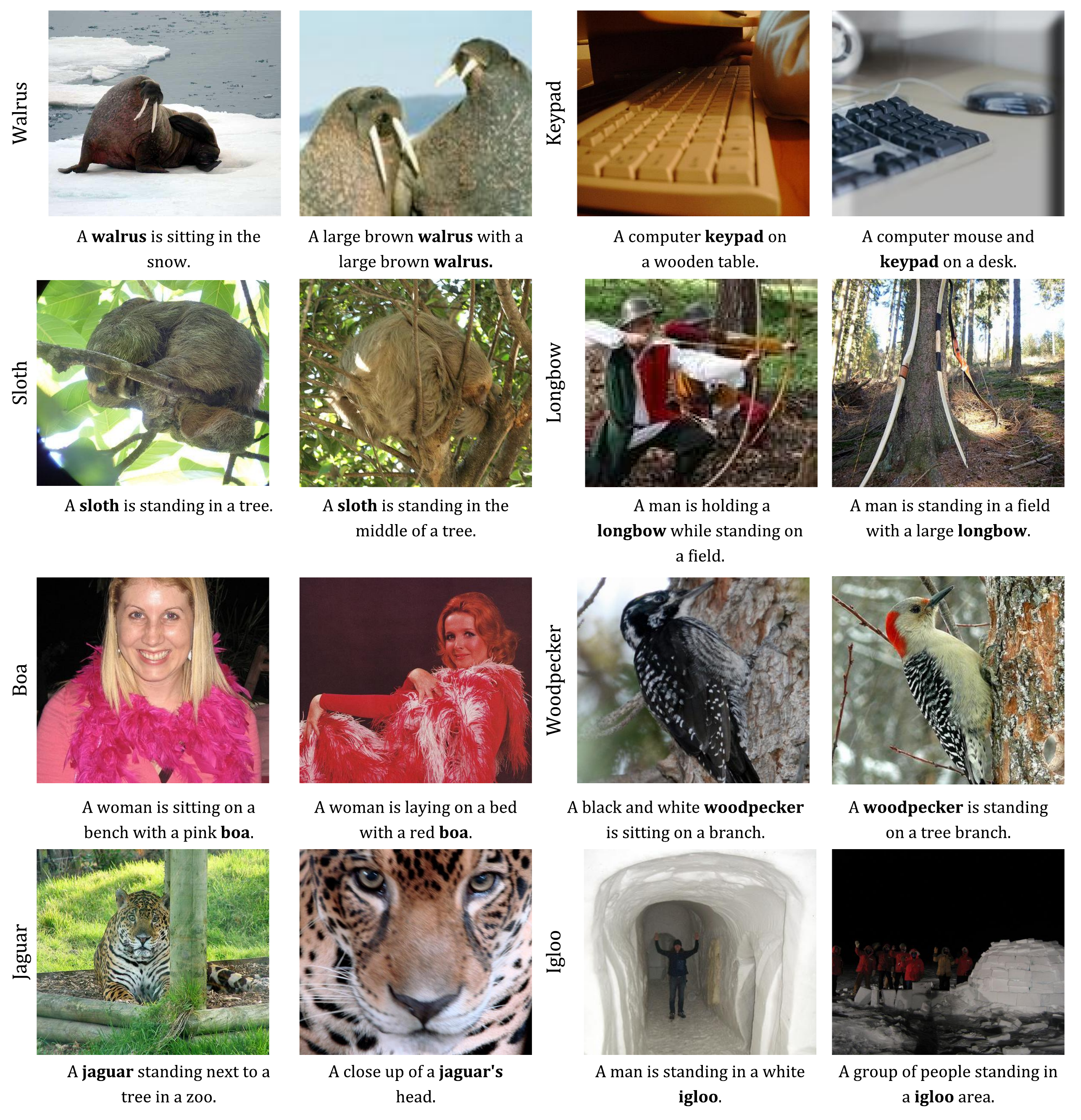}
\end{center}
    \caption{Image Description: Example \modelName{} descriptions for eight different objects.  \modelName{} is able to describe objects in different contexts.}
    \label{fig:many_descriptions3}
\end{figure*}

\paragraph{Successful Descriptions}
We highlight sentences generated by \modelName{} in Figure~\ref{fig:many_descriptions1}, \ref{fig:many_descriptions2}, and \ref{fig:many_descriptions3}.
By placing different images of the same object side-by-side, we can compare how the same object is described in different contexts.
For example, in Figure~\ref{fig:many_descriptions1}, a gecko is described as ``A person holding a gecko in their hand'' and ``A gecko is standing on the branch of a tree'' demonstrating that \modelName{} is able to describe a single object in a variety of ways to reflect different visual contexts.

\begin{figure*}
\vspace*{-0.25cm}
\begin{center}
\includegraphics[scale=0.6]{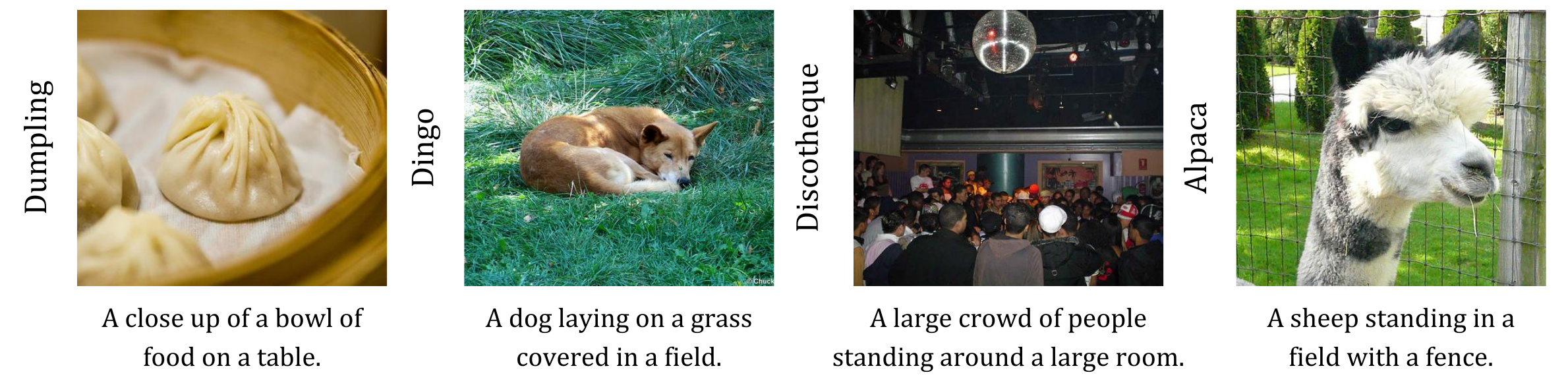}
\end{center}
    \caption{Image Description: Example descriptions generated by \modelName{} which are relevant, but do not describe a new object category.  Sometimes new objects are misclassified (e.g., ``sheep'' instead of ``alpaca'') but still describe the correct context.  Other times, the new object category is not needed to accurately describe the image.}
    \label{fig:error-no-new-object}
\end{figure*}

\paragraph{Errors}
We highlight common errors generated by \modelName{}.
Figure~\ref{fig:error-no-new-object} shows examples in which the description does not mention the new object category, but is still highly relevant.
Sometimes, \modelName{} produces accurate descriptions without mentioning the new object category.
Other times, \modelName{} correctly describes other elements of the scene, but misclassifies the new object category.  
For example, in Figure~\ref{fig:error-no-new-object} right, the model classifies the ``alpaca'' as a ``sheep''.

\begin{figure*}
\vspace*{-0.25cm}
\begin{center}
\includegraphics[scale=0.6]{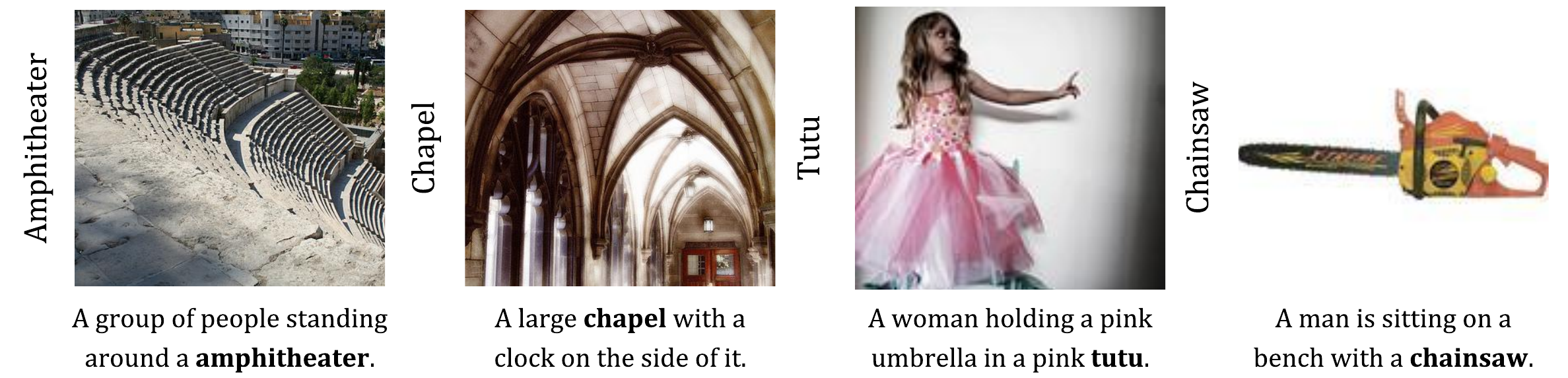}
\end{center}
    \caption{Image Description: Example descriptions generated by \modelName{} in which a new object is described, but the context is described incorrectly.  This is especially common in images, like the image of the chainsaw on the far right, which only include a single object on a monochromatic background.}
    \label{fig:error-out-of-context}
\end{figure*}

Figure~\ref{fig:error-out-of-context} provides examples of descriptions which do include a new object category, but describe it in the incorrect context.
DCC commonly hallucinates objects which are likely given the object context.
For example, when describing an amphitheater, DCC produces the sentence ``A group of people standing around a amphitheater'' even though no people are present in the image.
Such errors can be caused because the language model learns that people and amphitheaters occur together or because the lexical classifier mistakenly classifies ``people'' in the image.
When describing new types of buildings (e.g. chapel, fortress, or watchtower), DCC will frequently include the word ``clock'' in the description.  This is likely because MSCOCO vocabulary words like ``tower'' and ``building'' which are transferred to new words like ``chapel'' are frequently pictured with clocks.  
Sometimes the model includes objects in the description which are not contextually likely.
For example, when describing a little girl in a tutu, the model also describes an umbrella.  
This is likely because the visual features for ``umbrella'' overpower the language model.
Describing objects out of context is common for images which include a single object and a monochromatic background, as is commonly seen in ImageNet images.
Because only one object strongly activates the lexical layer, the model will frequently hallucinate objects which are not present in the image (e.g., ``A man is sitting on a bench with a chainsaw'' when there is no man or bench).

\begin{figure*}
\vspace*{-0.25cm}
\begin{center}
\includegraphics[scale=0.6]{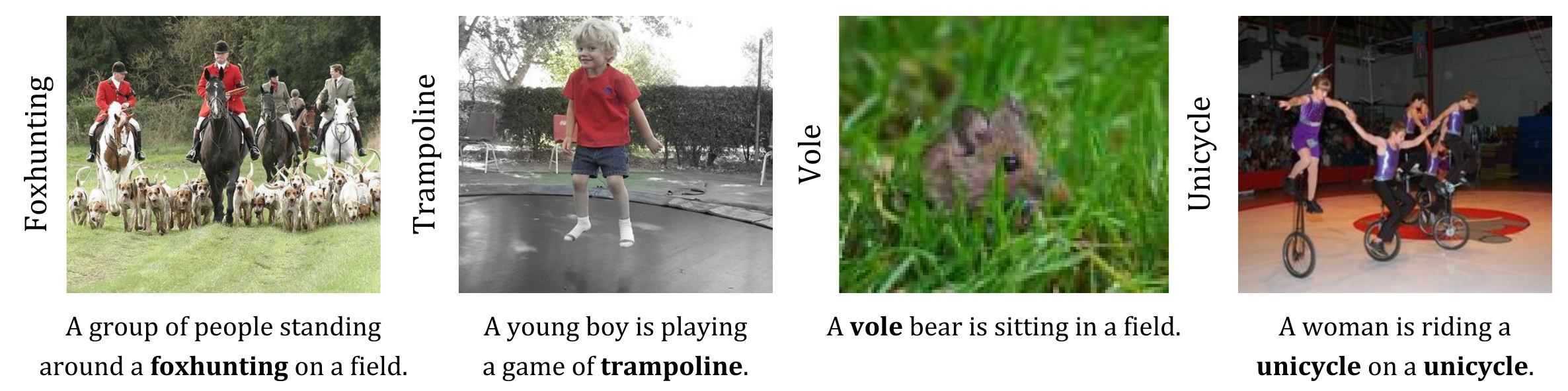}
\end{center}
    \caption{Image Description: Example descriptions generated by \modelName{} with poor grammar.  Poor grammar can be caused by a poor transfer word (the transfer words for ``foxhunting'' and ``trampoline'' are ``dog'' and ``frisbee'' respectively) or because the language model learns poor language features for the new object category.}
    \label{fig:error-grammar}
\end{figure*}

Figure~\ref{fig:error-grammar} provides examples of grammatical errors. 
As mentioned in the main paper, grammatical errors arise when a poor word is chosen for transfer.
Examples of poor grammatical sentences when the word ``dog'' is transferred to ``foxhunting'' and the word ``frisbee'' is transferred to ``trampoline'' are shown in Figure~\ref{fig:error-grammar}.
Another common grammatical mistake is for the model to list two objects in a row such as a ``vole bear'' or for the object to repeat a phrase such as ``a unicycle on a unicycle''. 
One possible explanation for such errors is that the pretrained language model does not learn good language features for these words.  
Consequently, after a new word is generated, the model generates a poor subsequent word.

\begin{figure*}
\vspace*{-0.25cm}
\begin{center}
\includegraphics[scale=0.6]{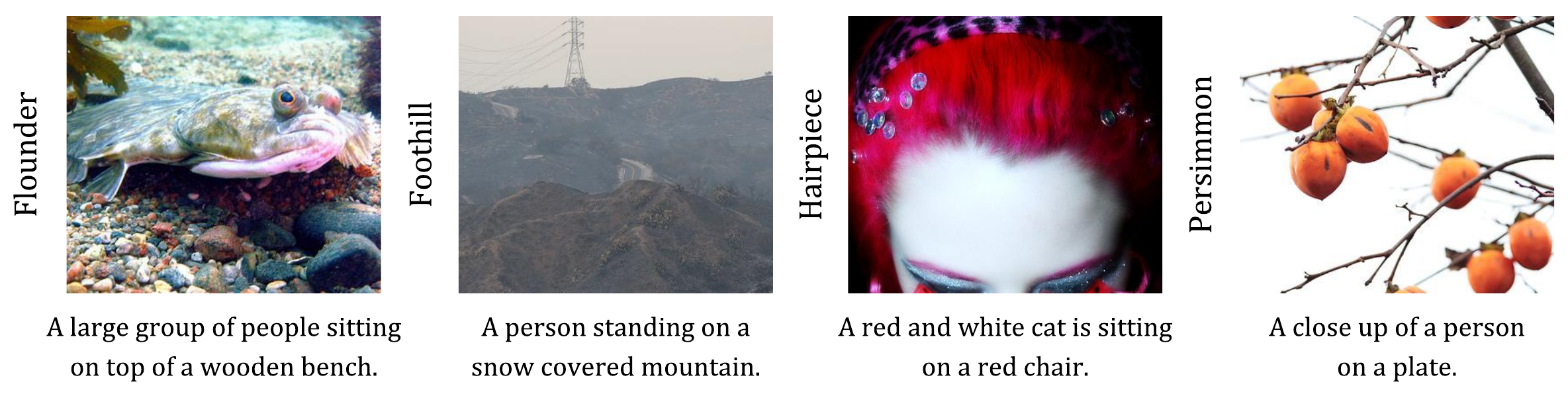}
\end{center}
    \caption{Image Description: Though most descriptions are relevant, some descriptions are incorrect.}
    \label{fig:error-irrelevant}
\end{figure*}

Finally, Figure~\ref{fig:error-irrelevant} shows images with irrelevant captions.

\section{Video Description}

\subsection{Empirical Results}

As was presented for image description in the main paper, we also explore the effects of training the language model used for video description with out-of-domain unpaired text data in Table~\ref{tab:yt}.
Comparing to Table 4 in the main paper, METEOR drops when training the language model with out-of-domain unpaired text (29.1 to 28.8 when including ImageNet videos during training and using transfer). The F1-score also drops when trained with ImageNet videos, but without training on ImageNet videos the F1-score actually increases from 6.0 to 13.3.  
\begin{table}[t]
\begin{center}
\begin{tabular}{|l|c|c|}
\hline
Model (Video, WebCorpus LM) & METEOR & F1 \\ \hline
Baseline (No Transfer) &  28.4 & 0.0 \\ 
 + DT &  28.4 & 13.3\\
 + ILSVRC Videos (No Transfer) &  28.8 & 0.0\\ 
 + ILSVRC Videos + DT &  28.8 & 13.7 \\ \hline
\end{tabular}
\end{center}
\caption{Video Description: METEOR scores across the test dataset and average F1 scores for the four held-out categories (All values in \%). The DCC models were trained on videos with 4 objects removed and the language model were trained on WebCorpus sentences.}
\label{tab:yt}
\end{table}
\subsection{Qualitative Results}

\begin{figure*}
\begin{center}
\includegraphics[scale=0.7]{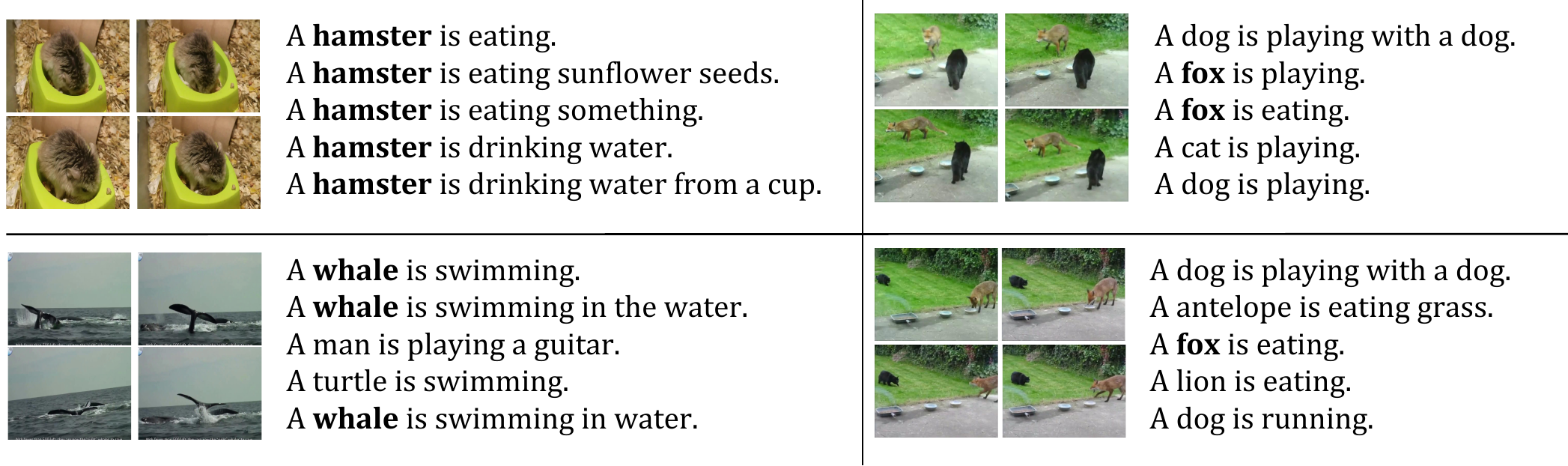}
\end{center}
    \caption{Video Description: Five most likely captions generated by DCC on novel objects unseen in paired training data.  Captions are sorted by likelihood with the top caption corresponding to the most likely caption.}
    \label{fig:qual_vid_imgnet}
\end{figure*}

We present qualitative results on ImageNet videos.  
The lexical classifier is trained with ImageNet and MSVD videos and the language model is trained with in-domain text data.
In addition to the objects held-out in the main paper (``zebra'', ``hamster'', ``broccoli'', and ``turtle'') we also describe videos which include the objects ``fox'' and ``whale''.
The caption model is never provided any paired video-sentence data which include these objects during training.
In  Figure~\ref{fig:qual_vid_imgnet}, we show the top five captions predicted by \modelName{} for videos which include ``whale'', ``fox'', and ``hamster''.
For ``whale'' and ``hamster'', \modelName{} predicts the correct object in the most probable caption.  
However, for ``fox'' the model predicts that ``dog'' is more probable than ``fox'', though ``fox'' is predicted in the second most probably caption.

\begin{figure*}
\begin{center}
\includegraphics[scale=0.9]{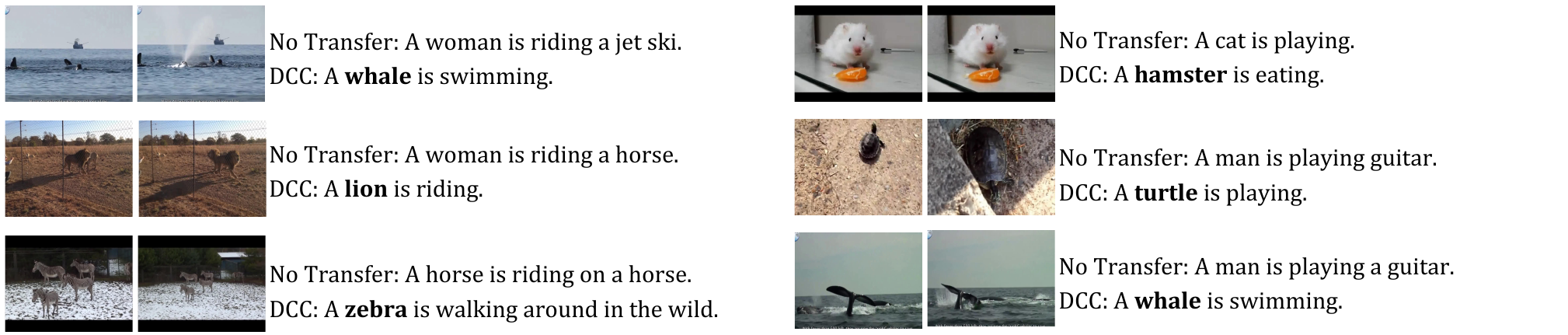}
\end{center}
    \caption{Video Description: Comparison of descriptions generated by \modelName{} for some of the objects in the ImageNet video dataset with and without transfer. Note that our model does not simply substitute added words in the place of words present in paired image-sentence data.}
    \label{fig:video_comparison_descriptions}
\end{figure*}

Figure~\ref{fig:video_comparison_descriptions} compares descriptions generated by the model without transfer and after transfer. The model is correctly able to identify and generate sentences to describe ``hamster'', ``lion'', ``turtle'', ``whale'', and ``zebra'' after transfer.
In comparison to the sentences produced by the model before transfer, the sentences generated after transfer describe the object and the context more accurately e.g. in the video containing a ``whale'', the sentence before transfer says ``A woman is riding a jet ski'' whereas after transfer \modelName{} says ``A whale is swimming'' which appropriately describes the object (whale) and the context (swimming) correctly.

Figure~\ref{fig:qual_vid_imgnet} also includes an example where the model has difficulty choosing the correct object. The five highest probability results each include a different animal in the description which indicates that the model is unsure which object is present.

\onecolumn
\twocolumn
\small
\bibliographystyle{ieee}
\bibliography{biblioShort,rohrbach,egbib,related,image2text}

\end{document}